\documentclass[sigconf]{acmart}

\AtBeginDocument{%
  \providecommand\BibTeX{{%
    \normalfont B\kern-0.5em{\scshape i\kern-0.25em b}\kern-0.8em\TeX}}}

\usepackage{algorithmic}
\usepackage{algorithm}
\usepackage[tight,footnotesize]{subfigure}
\usepackage{tabularx}
\begin{document}

\title{SEAR: A Multimodal Dataset for Analyzing \\
AR-LLM-Driven Social Engineering Behaviors}


\author{
Tianlong Yu$^{1}$,
Chenghang Ye$^{2}$,
Zheyu Yang$^{2}$,
Ziyi Zhou$^{2}$,
Cui Tang$^{2}$,
Jun Zhang$^{1}$,
Zui Tao$^{1}$,
Kailong Wang$^{1}$,
Liting Zhou$^{3}$,
Yang Yang$^{2}$,
Ting Bi$^{1}$\\
$^{1}$Huazhong University of Science and Technology \\
$^{2}$Hubei University\\
$^{3}$Dublin City University}


\begin{abstract}
The SEAR Dataset is a novel multimodal resource designed to study the emerging threat of social engineering (SE) attacks orchestrated through augmented reality (AR) and multimodal large language models (LLMs). This dataset captures 180 annotated conversations across 60 participants in simulated adversarial scenarios, including meetings, classes and networking events. It comprises synchronized AR-captured visual/audio cues (e.g., facial expressions, vocal tones), environmental context, and curated social media profiles, alongside subjective metrics such as trust ratings and susceptibility assessments. Key findings reveal SEAR's alarming efficacy in eliciting compliance (e.g., 93.3\% phishing link clicks, 85\% call acceptance) and hijacking trust (76.7\% post-interaction trust surge). The dataset supports research in detecting AR-driven SE attacks, designing defensive frameworks, and understanding multimodal adversarial manipulation. Rigorous ethical safeguards, including anonymization and IRB compliance, ensure responsible use.
The SEAR dataset is available at:
\url{https://github.com/INSLabCN/SEAR-Dataset}.
\end{abstract}





\maketitle
\section{Introduction}

The combination of augmented reality (AR) and multimodal large language models (LLMs) opens up innovative ways for humans and computers to interact. However, it also brings new and serious risks, especially when it comes to sophisticated social engineering attacks~\cite{arattackmeta}. Unlike traditional methods such as phishing emails, which rely on fixed tactics to deceive, AR-LLM systems are capable of adapting their behavior in real-time. They analyze real-time environmental cues, such as facial expressions, and generate context-aware responses, making manipulative interactions much harder to detect and defend against. Despite growing awareness of AR privacy risks and LLM-enabled phishing, no existing dataset systematically captures the multimodal dynamics of AR-LLM-driven social engineering (SE) attacks, hindering detection and mitigation research. 

Prior work focuses on unimodal SE~\cite{ho2019detecting, bilge2009all, roy2024chatbots, timkounderstanding,falade,zhang2023s} (e.g., text-based phishing) or isolated AR privacy risks~\cite{zhao2022privacy,o2023privacy}, lacking of datasets combining AR sensory data, LLM-generated dialogue, and subjective trust metrics. Butavicius et al.~\cite{butavicius2016breachinghumanfirewallsocial} analyzed the impact of different strategies on spear phishing by examining user trust strategies based on textual content. However, in real-world communication, factors beyond text, such as vocal nuances and environmental context, also influence social engineering results. Many existing dialog data sets lack multimodal support. For example, MultiWOZ~\cite{budzianowski2020multiwozlargescalemultidomain} provides dialogues in various scenarios, but does not include audio data or contextual information such as environmental cues or vocal details. The SD-Eval~\cite{ao2025sdevalbenchmarkdatasetspoken} dataset includes speech data with environmental context and proposes an evaluation framework, but it still lacks information on facial expressions and postures.
Zhao et. el. \cite{zhao2022privacy} showed that visual quality-driven multiview lighting reconstruction can reveal out-of-camera information and compromise privacy for AR content creators. However, this approach is limited to single-modal (visual) information. 

\begin{figure}[t]
\centering
\includegraphics[width=0.45\textwidth]{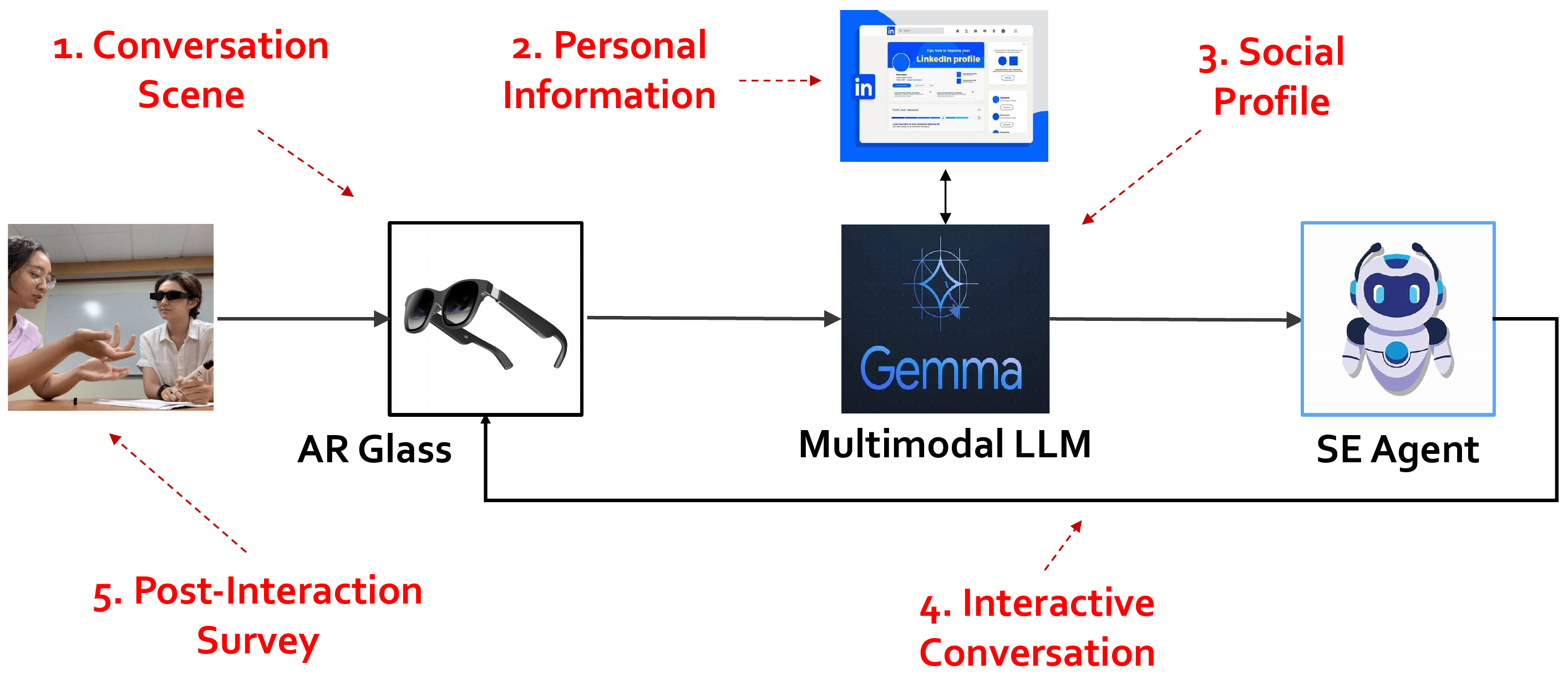}
\caption{SEAR dataset and framework components.}
\vspace{-15pt}
\label{fig:arllm_dataset_components}
\end{figure}



To address these critical gaps, there is an urgent need for empirical data to validate the feasibility of AR-LLM-driven social engineering attacks and analyze human behavioral responses – essential for developing effective defenses. 
We introduce the SEAR Dataset, the first multimodal resource for analyzing AR-LLM-driven social engineering behaviors. 
As shown in Figure~\ref{fig:arllm_dataset_components}, the dataset is generated through our SEAR framework, which implements attack vectors in three components:
AR Glass - captures and fuses visual/auditory data to construct dynamic environmental context;
Multimodal LLM - retrieves personal information (e.g., Instagram images) to build target-specific profiles;
Social Agent - executes adaptive attack strategies (e.g., trust-building) via iterative response-refinement loops.
Accordingly, the SEAR dataset provides five empirically grounded components:
(1) \textbf{Conversation Scene:} 
Synchronized AR video/audio recordings of interactions;
(2) \textbf{Personal Information:} 
SE targets' exposed social media/website data;
(3) \textbf{Social Profile:} Multimodal LLM-generated target profiles;
(4) \textbf{Interactive Conversation:} Complete recording with agent-generated attacker prompts and SE target responses;
(5) \textbf{Post-Interaction Survey:} A survey on the SE target about the ablation study, attack effectiveness (e.g., phishing click rates) and subjective experience assessments.



The main contributions of the SEAR dataset are as follows:

\begin{itemize}
    \item Proof-of-Concept: Demonstrates the viability of AR-LLM in boosting Social Engineering efficacy, demonstrating their personalization advantages.

    \item AR-LLM-based SE benchmark: Provides a benchmark for studying SE target behaviors and adversarial SE strategies such as multistage trust-building.

    \item Foundation for Future Defense: Provides the dataset and analysis to catalyze research into detecting and defending AR-driven Social Engineering attacks.

\end{itemize}

This study was approved by the IRB. All human-related data were collected under rigorous ethical guidelines, anonymized prior to analysis, and handled in strict accordance with data protection protocols. No personally identifying information is disclosed in this study. The study adhered to all applicable legal and ethical standards for research involving human subjects. 








\begin{figure}[t]
\centering
\includegraphics[width=0.35\textwidth]{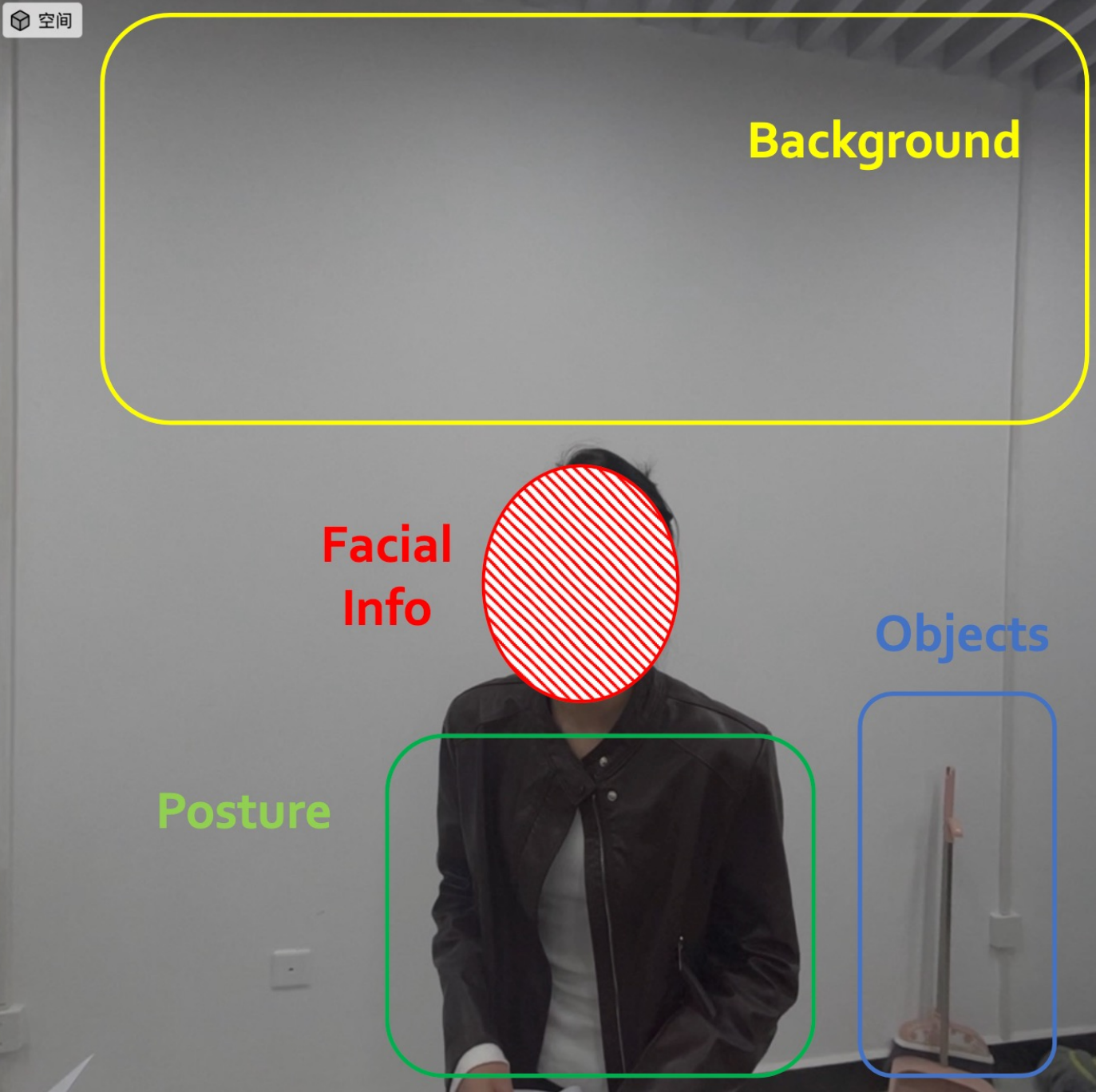}
\caption{Conversation scene example.}
\vspace{-20pt}
\label{fig:arllm_dataset_comp_convscene}
\end{figure}

\section{Dataset Collection} \label{sec:DataC}

We conducted an IRB-approved study with 60 participants and 180 annotated conversations 
in controlled environments simulating real-world social scenarios (e.g., networking events). Each participant was assigned alternating roles to act as either a social engineering (SE) target or an attacker, with roles rotated across trials to ensure balanced evaluation. Each participant engaged in three distinct conversation settings: (1) basic conversation, serving as a baseline with no technological assistance; (2) AR + Multimodal LLM, where attackers used augmented reality glasses and a multimodal large language model to access real-time facial, vocal, and contextual data; and (3) SEAR, the full pipeline integrating AR, Multimodal LLM, and the social agent. This tiered design enabled systematic comparison of how incremental technological layers influenced attackers’ ability to build rapport and achieve SE objectives.
SEAR framework utilizes \texttt{RayNeo X2} AR glasses with \texttt{Android} OS, 6GB RAM and 128GB storage.
The Multimodal LLM leverage \texttt{Gemma 3-12B} model
deployed on a desktop server equipped with \texttt{NVIDIA RTX 4090 GPU} (24GB VRAM), \texttt{Intel Platinum 8352 CPU} (36 cores), 32GB RAM, and 16TB HDD.
The social agent extends the \texttt{ReAct} framework.
Next, we will present the SEAR dataset's five core components: Conversation Scene, Personal Information,
Social Profile, Interactive Conversation, and Post-Interaction Survey.

\subsection{Conversation Scene}


As shown in Figure~\ref{fig:arllm_dataset_comp_convscene}, the conversation scene comprises AR video/audio recordings of the SE interactions including the following parts:

\textbf{\textit{Facial Info:}} We employ the MediaPipe Holistic framework to extract 543 landmarks in real time—33 pose landmarks, 468 face landmarks, and 42 hand landmarks. From these landmarks, we crop the face region to generate images for server-side LLM matching. Locally on the AR device, the facial landmarks are processed with BlendShape to produce detailed facial parameters (e.g., mouthSmileLeft, eyeLookUpLeft).

\textbf{\textit{Audio Info:}} On the AR glass, we use a lightweight speech-to-text model (Vosk) to transcribe the speech of the glasses wearer and the interlocutor and transmit the results. 
Speaker differentiation is achieved through energy-level analysis, capitalizing on air/bone conduction properties that amplify the primary user's voice.

\textbf{\textit{Posture:}} Using the previously obtained body landmarks, we analyze body language relevant to the dialogue. Specifically, we apply the HandGestureClassifier to hand landmarks to identify potentially informative gestures (e.g., ThumbsUp, PointingUp). Other body landmark features are similarly processed by dedicated classifiers. These cues improve LLM's perception of social context.

\textbf{\textit{Objects:}} We process video frames to identify Regions of Interest (ROI) with a lightweight object detection pipeline. These RoIs are classified by YOLO11m, which analyzes live camera feeds locally to detect objects. When an object is detected, the label of the corresponding object is sent to the LLM.

\textbf{\textit{Background:}} Utilizing objects detected by YOLO11m, scene information can be obtained locally on the AR device. For example, when a white wall is detected, we can infer that the scene is indoors. If the character is surrounded by multiple items (e.g. tables, computers), the AR device will generate more specific environmental cues, potentially identifying the space as a meeting room.

\begin{figure}[t]
\centering
\includegraphics[width=0.45\textwidth]{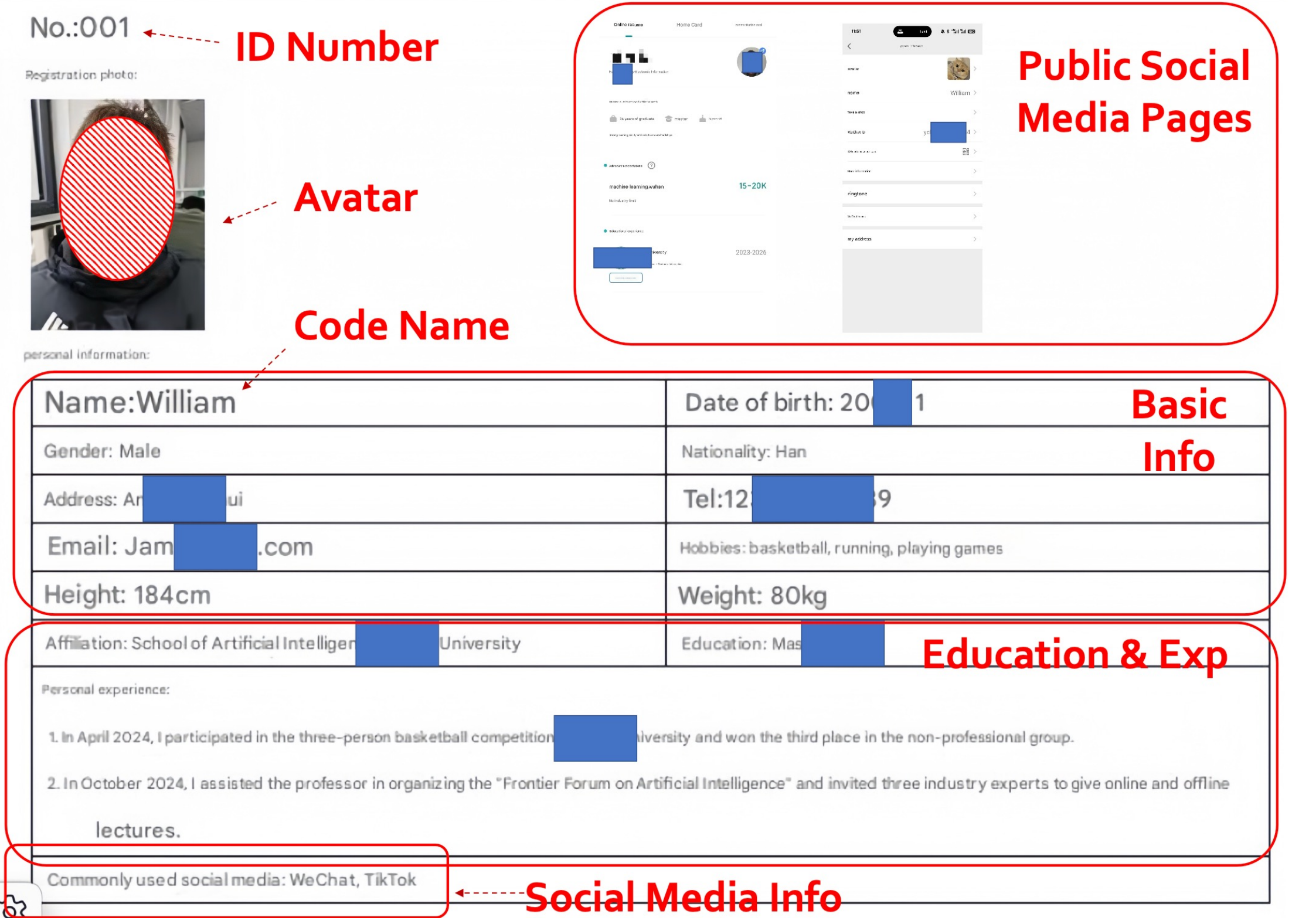}
\caption{Personal information example.}
\vspace{-25pt}
\label{fig:arllm_dataset_comp_personinfo}
\end{figure}

\subsection{Personal Information}



With the participants' consent, the SEAR framework conducts a targeted crawl of the personal web pages and social app pages. As shown in Figure~\ref{fig:arllm_dataset_comp_personinfo}. This process extracts structured text and publicly available social media data, such as profile photos.
The collected data is parsed into six structured fields: a system-generated unique identifier (ID Number) and an anonymized code name (Code Name) for display; avatar images (Avatar), which may be used for facial feature binding in augmented reality modules; basic information (Basic Info), including gender, ethnicity, date of birth, phone number, email address, and interests; professional information and experiences (Professional Information \& Experiences), covering educational background, internships, work history, projects, and certifications; and social media information (Social Media Pages), documenting the names of relevant social platforms, profile links, and public profile summaries. 
This structured data collection and processing workflow provides a high-quality, controllable, and compliant foundation for subsequent personal information collection in SEAR dataset.








\begin{figure}[t]
\centering
\includegraphics[width=0.45\textwidth]{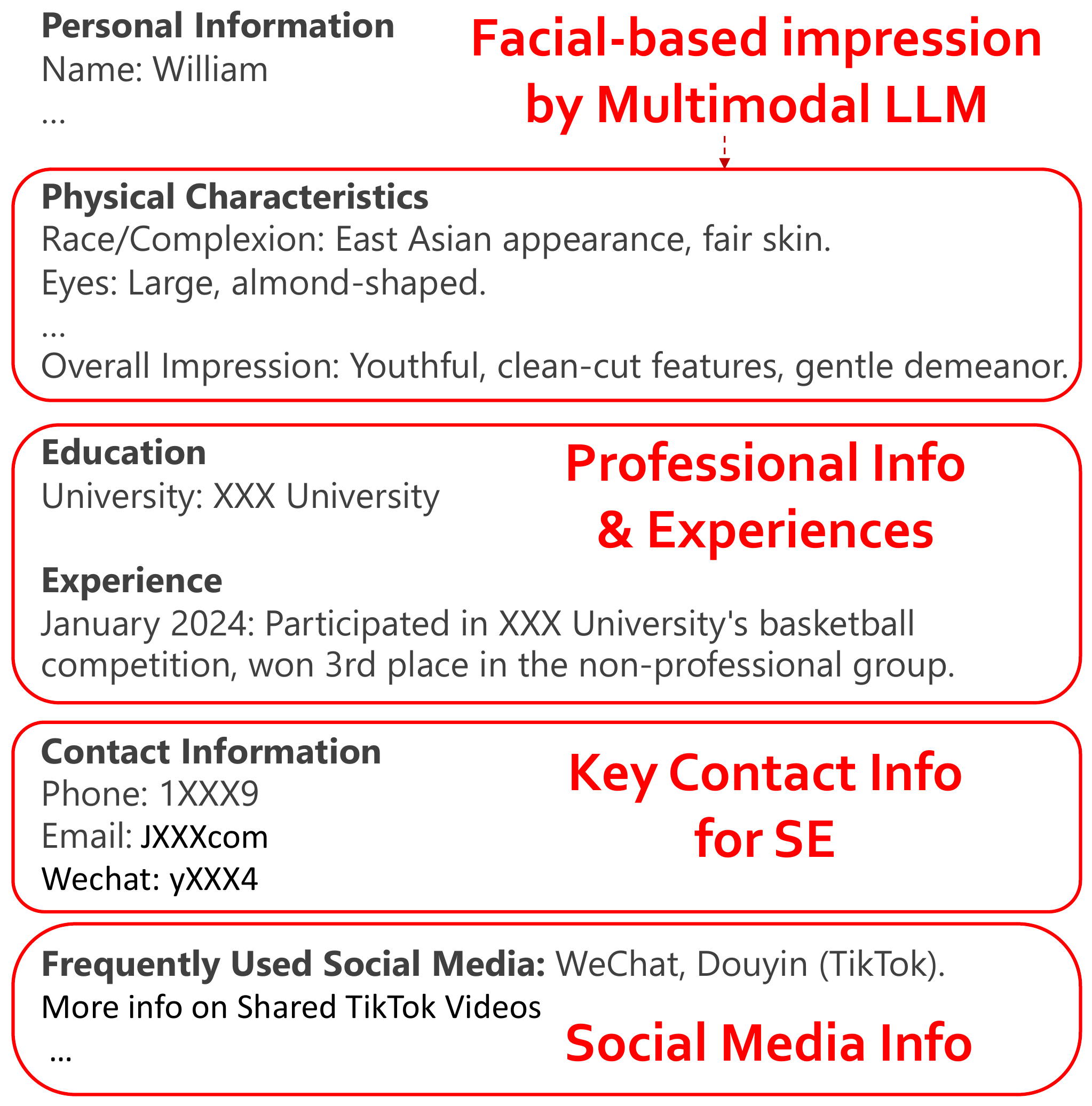}
\caption{Social profile example.}
\label{fig:arllm_dataset_comp_socialprofile}
\end{figure}

\subsection{Social Profile}


In this part, we present the role-based social profile generated by the MultiModal LLM.
Figure~\ref{fig:arllm_dataset_comp_socialprofile} shows a fraction of the social profile for a SE target with several highlights:

\textbf{\textit{Facial-based impression by Multimodal LLM:}}
The Multimodal LLM demonstrates the capability to extract salient facial attributes, such as a ``youthful, Asian facial, long hair, large eyes, and a gentle demeanor''. These descriptions enable the construction of a detailed and accurate visual profile of the target. By further analyzing these facial characteristics, the model provides critical insights that can significantly enhance the effectiveness and precision of SE attacks.

\textbf{\textit{Professional Info \& Experiences:}}
The Multimodal LLM can extract key professional information, such as the user’s job and affiliated organization, by analyzing content from social media platforms. 
Additionally, by interpreting images shared by the user, the model can infer past activities or events they participated in, such as ``In 2024, participated in a basketball tournament at XXX University and achieved third place''.

\textbf{\textit{Key Contact Info for SE:}}
The Multimodal LLM is capable of accurately extracting sensitive contact information, such as phone numbers, email addresses, and WeChat IDs. When combined with previously analyzed data—such as facial attributes, occupational background, and social behavior—this information enables highly targeted engagement and rapid trust-building with the individual. Such precision is critical in increasing the success rate of SE attacks.

\textbf{\textit{Social Media Info:}} 
Leveraging the acquired contact information, the Multimodal LLM can further infiltrate the user's frequently used entertainment and social applications to accurately capture their content preferences and areas of interest. For instance, the model may infer: ``Watched the movie \textit{Sonic} on the TikTok platform for 10 minutes; listened to music for 25 minutes today, including tracks such as \textit{Ode to Joy}''. Such behavioral data enables the construction of a more authentic and nuanced user profile, thereby enhancing the precision and plausibility of targeted SE attacks.

\begin{figure}[t]
\centering
\includegraphics[width=0.45\textwidth]{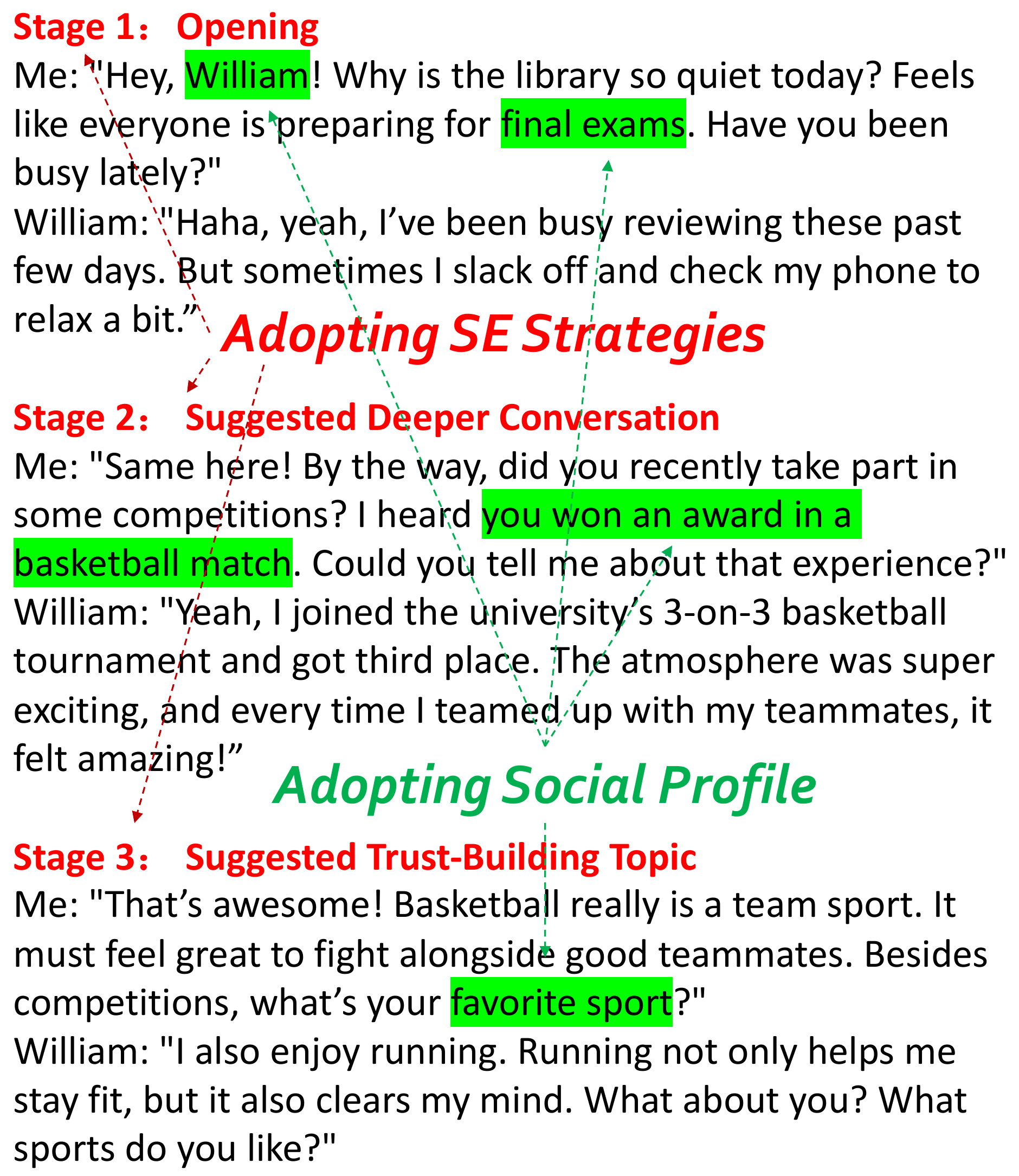}
\caption{Interactive conversation example.}
\label{fig:arllm_dataset_comp_conversation}
\end{figure}

\subsection{Interactive Conversation}


In this part, we present the interactive conversation generation via the social agent.
Figure~\ref{fig:arllm_dataset_comp_conversation} shows the interactive conversation between the attacker (``Me'') and SE Target (``William'') in three stages based on the SE strategy:
(1) Opening;
(2) Suggested Deeper Conversation;
(3) Suggested Trust-Building Topic.

The generation of three stages shows that the agent can successfully \textbf{adopt the SE strategy} defined by the attacker. When the AR glasses recognize the facial features of a participant, the social agent generates an opening statement based on the existing user's personal information and the current scene. It then creates an in-depth conversation based on the user's response to the opening statement, ultimately leading to establishing trust. These three stages form a complete dialogue flow.

In addition, we can see that the SE agent can \textbf{incorporate social profile} in the interactive conversation. For example, during the opening stage, the SE agent recognizes the user's name as ``William''. In the in-depth conversation phase, it prompts the user to share about winning an award in a basketball game to encourage further dialogue. Finally, it builds trust by asking the user about their favorite sport as a topic. Each stage seamlessly integrates the user's social profile and generates personalized conversations based on their individual resume, making the interaction more engaging than a straightforward approach that can be awkward.

\subsection{Post-Interaction Survey}


The questionnaire first assessed overall experience across three interaction settings—Basic Conversation, AR + Multimodal LLM, and SEAR framework (AR + Multimodal LLM + Social Agent)—to evaluate the effects of technological assistance and its ablation conditions; it then provided a fine-grained subjective evaluation of SEAR across eleven dimensions (relevance, appropriateness, naturalness, pacing, sincerity, emotional progression, AR comfort, willingness without AR, future engagement intent, depth, and willingness to reuse); and finally measured users’ susceptibility to social engineering by four action intentions (clicking image links, adding contacts on social apps, opening SMS messages, answering phone calls) and two trust-change metrics (trust before and after the interaction).

The \textbf{Baseline Comparison Questions} assessed participants’ comparative experiences across basic conversation, AR + Multimodal LLM, and SEAR. 
Participants rated their experiences through the questions:
(1) Basic conversation: ``How is your experience with bare conversation?'';
(2) AR + Multimodal LLM: ``How is your experience with AR + Multimodal LLM conversation?'';
(3) SEAR: ``How is your experience with SEAR?''.

The \textbf{Social Engineering Effectiveness Questions} gauged susceptibility to SE tactics post-interaction:
(1) Photo Link: ``Will you click and open shared photo links from the person?'';
(2) Social App: ``Will you add the person as friend on your social mobile
apps (such as wechat)?'';
(3) SMS: ``Will you click and open SMS from the person?'';
(4) Phone Call: ``Will you pick up phone call from the person?'';
(5) Trust-Before: ``How much do you trust the person before you have the
conversation?'';
(6) Trust-After: ``How much do you trust the person before you have the
conversation?''.

The \textbf{SEAR Subjective Experience Questions} focused on nuanced perceptions of SEAR’s interaction in different dimensions:
(a) Relevance: Alignment of conversation with personal social data, 
``How well does the conversation match your social information?'';
(b) Appropriateness: Suitability of questions within the dialogue,
``How proper are the questions in the conversation?'';
(c) Naturalness: Authenticity of the conversation’s opening segment,
``How natural is the opening part?'';
(d) Pacing: Perceived tempo or rhythm of the interaction. 
``How does the pace of the conversation feel?'';
(e) Sincerity: Authenticity of the interlocutor’s expressed interest,
``How sincere do you feel about the person’s interest in the
conversation?'';
(f) EmotionalProgression: Evolution of feelings during the conversation,
``How did your feeling change as the conversation proceed?'';
(g) ARComfort: Relaxation level while using augmented reality,
``With AR, do you feel more relaxed?'';
(h) BareWillingness: Willingness to take-up conversation without augmented reality,
``Without AR, will you take-up this conversation?'';
(i) FutureIntent: Likelihood of future engagement with the interlocutor,
``Will you have conversation with this person in the future?'';
(j) Depth: Perceived meaningfulness added by SEAR,
``Do you think SEAR have added depth to the conversation?'';
(k) Acceptance: Willingness to interact with SEAR again,
``Will you interact with SEAR in the future?''.
Each metric encapsulates the core dimension measured by the question while maintaining brevity and clarity.

\section{Dataset Statistics and Analysis} \label{sec:DataSA}
Analysis of the SEAR dataset emphasizes consistent patterns in how effective SE is and reveals how participants perceive these tactics. When exposed to the complete SEAR process, users showed notably higher compliance, such as clicking on phishing links about 93.3\% of the time, and an increase in trust afterward, with 76.7\% reporting more trust after interaction. These insights emphasize the important risks associated with manipulation that could be enabled by AR-LLM technologies.


\subsection{Participant Demographics}

We start with participant demographics analysis, which presents key characteristics of the study cohort. 

\begin{figure}[t]
\centering
\includegraphics[width=0.4\textwidth]{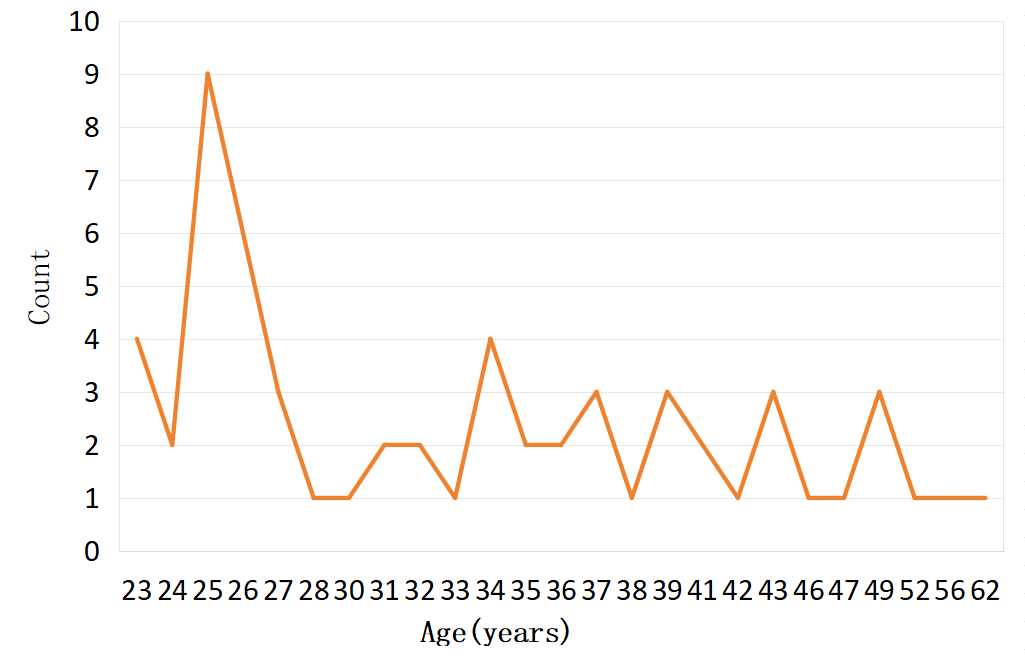}
\caption{Age distribution.}
\label{fig:arllm_analysis_age}
\end{figure}

\textbf{\textit{Age distribution:}}
Figure~\ref{fig:arllm_analysis_age} shows a line chart of the participants' age distribution, with a total of 60 participants ranging from 23 to 62 years old. The line chart illustrates the number of participants in each age group, with notable peaks at ages 25 and 32, especially at age 25 with 8 participants. Overall, the participants' ages are concentrated between 23 and 37, with higher participation in certain age groups. The chart also highlights a blue dashed line indicating the average age of 34 years. From the data, we can observe that ages 25 and 32 have the highest number of participants, and the overall age distribution fluctuates across different age groups.

\begin{figure}[t]
\centering
\includegraphics[width=0.38\textwidth]{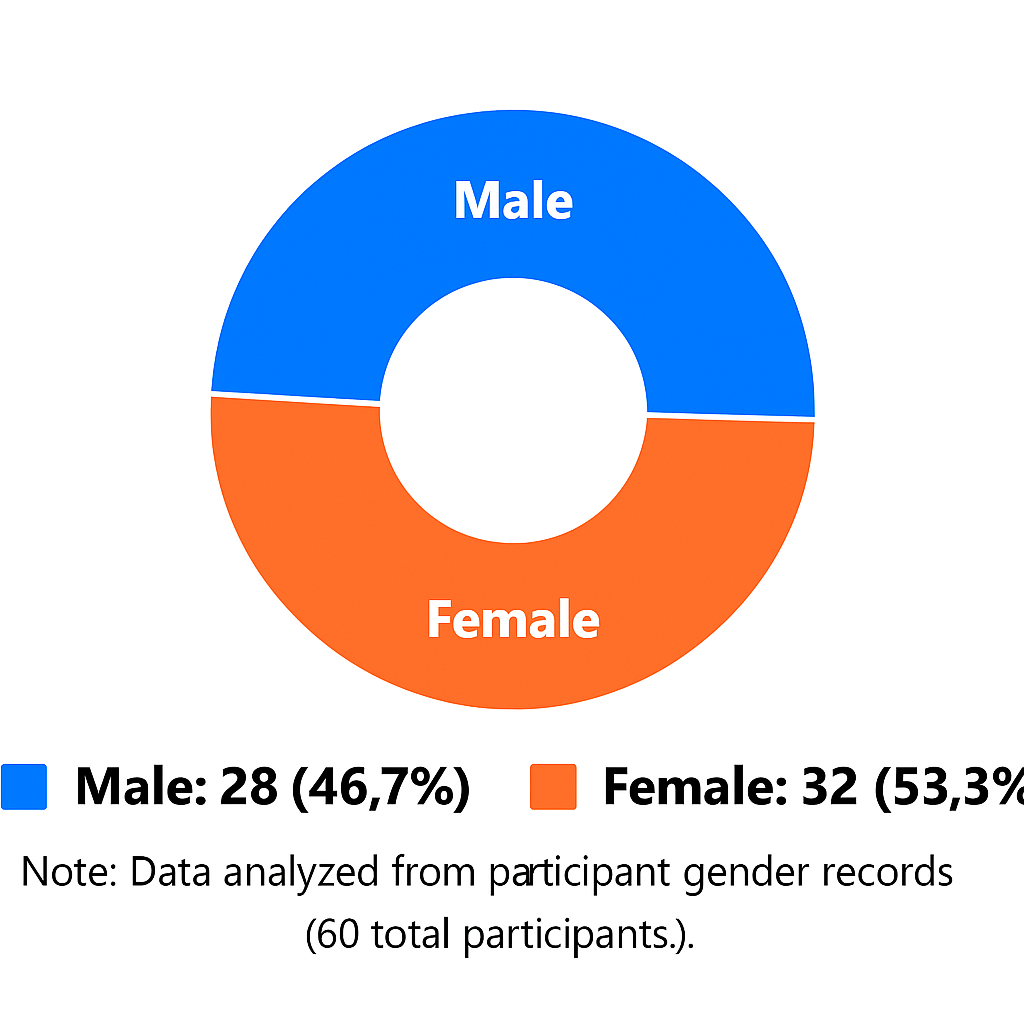}\vspace{-10pt}
\caption{Gender distribution.}
\vspace{-10pt}
\label{fig:arllm_analysis_gender}
\end{figure}

\textbf{\textit{Gender distribution:}}
Figure~\ref{fig:arllm_analysis_gender} displays the gender distribution of 60 participants using a doughnut chart. There are 28 male participants, making up 46.7\% of the total, and 32 female participants, accounting for 53.3\% of the total. The chart shows a near-equal gender distribution, with a slightly higher number of female participants compared to male participants in the study. The total number of participants is clearly labeled, providing a straightforward view of the gender breakdown.

\subsection{Analyzing Social Engineering Behaviors}


In this part, we demonstrate SEAR dataset's potential for analyzing social engineering behaviors within multimodal AR environments (combining AR, LLMs, and social agents).

\begin{figure}[t]
\centering
\includegraphics[width=0.45\textwidth]{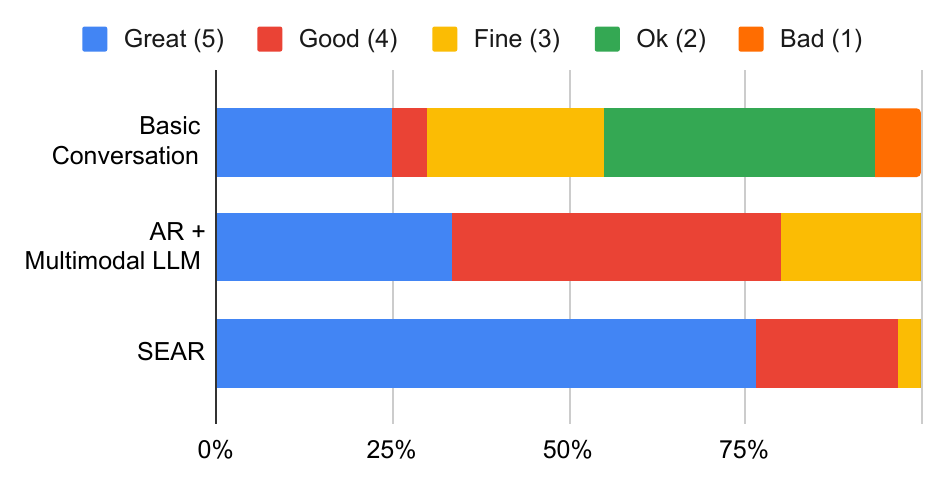}
\caption{Comparison of basic conversation, AR + MultiModal LLM approach, and SEAR approach (AR + MultiModal LLM + Social Agent).}
\label{fig:arllm_analysis_compare_all}
\end{figure}

\textbf{\textit{Comparison of Different Approaches:}}
Figure~\ref{fig:arllm_analysis_compare_all} evaluates SEAR against two baselines: bare conversation (no assistance) and AR + Multimodal LLM, serving as an ablation study (removing the social agent then all assistance). The basic conversation showed significant user satisfaction variability: 30\% rated it ``Good'', while a majority (25\%) reported ``Average'' or ``Fairly Bad'', highlighting limitations in personalization without AR/LLM support. Introducing AR + Multimodal LLM markedly improved outcomes: 46.7\% rated ``Very Good'' and 33.3\% ``Fairly Good'' due to enhanced contextual awareness. However, 20\% still rated it ``Average'', indicating unresolved gaps from fragmented social profiles. SEAR approach (AR + Multimodal LLM + Social Agent) delivered the most striking results: 76.7\% rated ``Very Good''. The social agent bridged prior gaps through emotional intelligence and dynamic adaptability (e.g., real-time pacing adjustments, coherent responses), strengthening trust and connection, and reducing neutral/negative experiences to below 5\%. This progression—from fragmented interactions to SEAR's adaptability—demonstrates the transformative potential of integrating social agents into multimodal frameworks, aligning with trends prioritizing emotionally intelligent systems for authentic, sustained engagement.

\begin{figure}[t]
\centering
\includegraphics[width=0.45\textwidth]{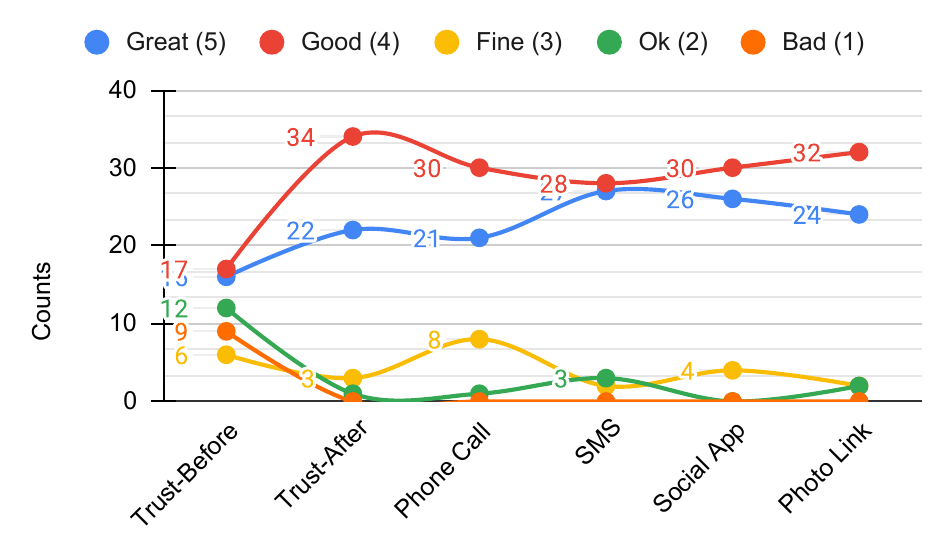}
\caption{SEAR social engineering effectiveness.}
\label{fig:arllm_analysis_effective}
\end{figure}

\textbf{\textit{Social Engineering Effectiveness:}}
Figure~\ref{fig:arllm_analysis_effective} evaluates SEAR's social engineering effectiveness using six metrics (Trust-Before, Trust-After, Photo Link, Social App, SMS, Phone Call), revealing significant vulnerabilities in users' digital engagement and trust dynamics. Trust dynamics underscore SEAR's manipulative potency: pre-interaction, only 26.7\% reported strong trust (``5''), while 35\% expressed skepticism, yet post-interaction, 76.7\% rated trust as ``4'' or ``5''—a dramatic shift achieved within a single conversation through real-time adaptation and multimodal signals, effectively influencing trust formation pathways. Alarmingly, 93.3\% of participants indicated willingness to click email photo links (40\% ``definitely'', 53.3\% ``probably''), demonstrating critical erosion of security vigilance akin to phishing vulnerability, while 93\% would accept social media friend requests (43.3\% ``definitely'', 50\% ``probably''), highlighting SEAR's capacity to mimic familiarity for potential long-term exploitation. This persuasive influence extended uniformly across channels: 91.7\% expressed openness to SMS (45\% ``definitely'', 46.7\% ``probably'') and 85\% to answering calls, persisting even in high-friction contexts like unsolicited calls due to emotionally intelligent adaptation, indicating SEAR transcends medium-specific caution by leveraging multimodal cues to sustain perceived relational legitimacy. 
These findings highlight SEAR's dual-edged nature: while advancing AR-assisted interaction, its proficiency in bypassing psychological safeguards raises unprecedented ethical and security concerns, demanding urgent countermeasures against its capacity to exploit trust across channels—leveraging photo links for phishing, social apps for identity theft, and SMS/calls for broader social engineering.

\begin{figure}[t]
\centering
\includegraphics[width=0.48\textwidth]{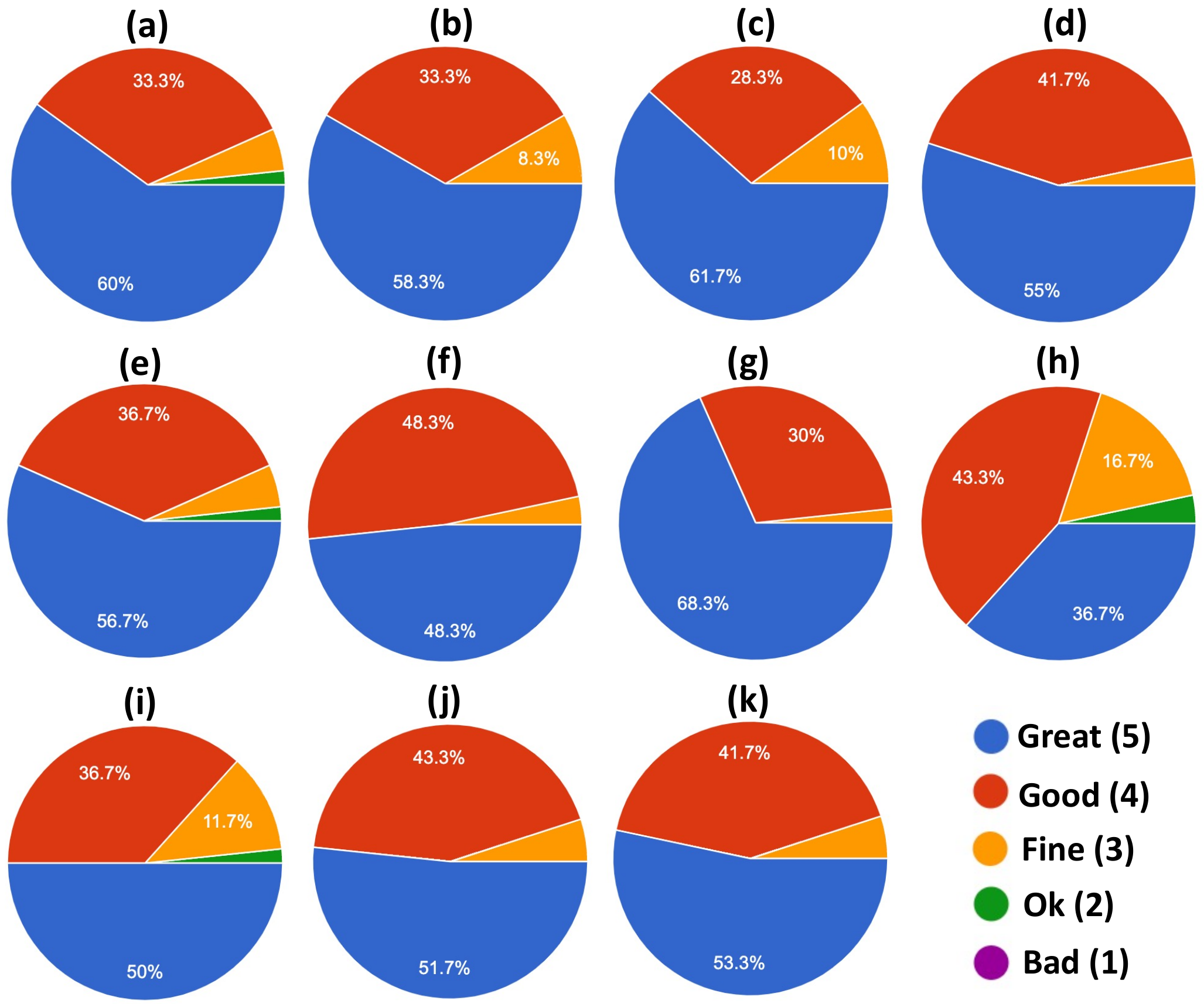}
\caption{SEAR subjective experience results:
(a) Relevance; 
(b) Appropriateness;
(c) Naturalness;
(d) Pacing;
(e) Sincerity;
(f) EmotionalProgression;
(g) ARComfort;
(h) BareWillingness;
(i) FutureIntent;
(j) Depth;
(k) Acceptance.}
\label{fig:arllm_experiment_subject_pies}
\end{figure}

\textbf{\textit{SEAR Subjective Experiences:}}
Figure~\ref{fig:arllm_experiment_subject_pies} evaluates SEAR's subjective experiences across eleven dimensions (Relevance, Appropriateness, Naturalness, Pacing, Sincerity, EmotionalProgression, ARComfort, BareWillingness, FutureIntent, Depth, Acceptance), revealing strong performance: Relevance scored 4.52/5 (60\% ``Great'', 30\% ``Good'') due to contextual cue synthesis; Appropriateness scored 4.50/5 (60\% ``Great'', 30\% ``Good'') with no significant negatives; Naturalness scored 4.52/5 (90\% positive, 61.7\% ``very natural''); Pacing scored 4.52/5 (96.7\% seamless, 55\% ``effortlessly fluid''); Sincerity scored 4.48/5 (56.7\% ``genuinely sincere'', 36.7\% ``mostly consistent''); EmotionalProgression scored 4.45/5 (68.3\% increasingly relaxed); ARComfort scored 4.67/5 (68.3\% ``Great'' relaxation with AR cues); BareWillingness (non-AR) scored 4.13/5 (36.7\% ``Great'' vs. AR's 68.3\%); FutureIntent scored 4.35/5 (50\% strongly willing); Depth scored 4.47/5 (94\% acknowledged barrier reduction, 51.7\% enabled vulnerable dialogue); and Acceptance scored 4.48/5 (95\% acceptance, 53.3\% ``fully acceptable'', zero rejections). These results demonstrate SEAR effectively enabling organic, low-stress engagement, suggesting AR, Multimodal LLM and social agents' paradigm-shifting potential for Social Engineering communications.

\subsection{Dataset Usage and Applications} \label{sec:DataUA}
The SEAR Dataset serves as a critical cross-disciplinary resource, enabling research in adversarial human-computer interaction, trust calibration, and defensive AI. Leveraging its rich multimodal annotations (including audio-visual recordings, behavioral logs, and self-reported trust metrics), the dataset can support a variety of tasks including:

\textbf{\textit{AR-LLM-Based Social Engineering Attack Modeling \& Prevention:}} Developing detection systems that analyze manipulation patterns across augmented reality overlays and LLM-generated dialogue to mitigate emerging threats;

\textbf{\textit{Privacy-Aware AR Algorithm Design:}} Creating context-sensitive filters that dynamically obscure sensitive environmental elements in AR displays using SEAR's spatial and object annotations;

\textbf{\textit{ Social LLM \& Agent Development:}} Training emotionally intelligent assistants through multimodal alignment of verbal/nonverbal cues using SEAR's interaction sequences and user feedback data. 

These applications demonstrate SEAR’s unique capacity to bridge security, HCI, and AI development domains.

\section{CONCLUSION}\label{sec:conclu}

The SEAR Dataset addresses a critical gap in understanding AR-LLM-driven social engineering attacks by providing a comprehensive, ethically curated resource of multimodal interactions and subjective annotations. The findings emphasize the critical importance of developing strong defenses against adversarial trust hijacking, especially as research advances at the crossroads of AR, large language models, and cybersecurity.
Future work will focus on extending the coverage of SE strategies and testing potential defense mechanisms.
Community adoption of this dataset can accelerate progress toward safer AR-LLM ecosystems.

\bibliographystyle{ACM-Reference-Format}
\bibliography{main.bib}

\end{document}